\documentclass[runningheads]{llncs}
\usepackage{graphicx}
\usepackage{amsmath,amssymb}
\usepackage{cite}
\usepackage{amsfonts}
\usepackage{algorithmic}
\usepackage{textcomp}
\usepackage{url}
\usepackage{float}
\usepackage{tabularx}
\usepackage{upquote}
\usepackage{color}
\usepackage{hhline}
\usepackage{mathtools}
\begin{document}

\title{Unsupervised RGBD Video Object Segmentation Using GANs} 
\titlerunning{ForeGAN\_RGBD} 


\author{Maryam Sultana\inst{1}\orcidID{0000-0002-8831-843X} \and
Arif Mahmood\inst{2}\orcidID{0000-0001-5986-9876} \and
Sajid Javed\inst{3} \and
Soon Ki Jung\inst{1}\orcidID{0000-0003-0239-6785}}
%

\authorrunning{M. Sultana et al.} 


\institute{School of Computer Science and Engineering, Kyungpook National University, Daegu, Republic of Korea, \\
\email{skjung@knu.ac.kr, maryam@vr.knu.ac.kr}\\
\and
 Department of Computer Science, Information Technology University (ITU), Lahore, Pakistan,\\
\email{arif.mahmood@itu.edu.pk}
\and
Department of Computer Science, University of Warwick, United Kingdom,\\
\email{s.javed.1@warwick.ac.uk}}

\maketitle

\begin{abstract}
Video object segmentation is a fundamental step in many advanced vision applications. Most existing algorithms are based on handcrafted features such as HOG, super-pixel segmentation or texture-based techniques, while recently deep features have been found to be more efficient. Existing algorithms observe performance degradation in the presence of challenges such as illumination variations, shadows, and color camouflage. To handle these challenges we propose a fusion based moving object segmentation algorithm which exploits color as well as depth information using GAN to achieve more accuracy. Our goal is to segment moving objects in the presence of challenging background scenes, in real environments. To address this problem, GAN is trained in an unsupervised manner on color and depth information independently with challenging video sequences. During testing, the trained GAN generates backgrounds similar to that in the test sample. The generated background samples are then compared with the test sample to segment moving objects. The final result is computed by fusion of object boundaries in both modalities, RGB and the depth. The comparison of our proposed algorithm with five state-of-the-art methods on publicly available dataset has shown the strength of our algorithm for moving object segmentation in videos in the presence of challenging real scenarios. 

\end{abstract}
\section{Introduction}
A fundamental step in many computer vision and artificial intelligence applications involves objects segmentation, for various tasks such as object detection \cite{wu2017moving, bouwmans2018applications}, visual object tracking \cite{zhang2015robust, vaswani2017robust}, video surveillance \cite{bouwmans2014robust, javed2016spatiotemporal}, salient motion detection \cite{chen2018scom} and image inpainting \cite{sultana2018unsupervised}. For object segmentation, background modeling plays a crucial role as it is a key process, which describes a scene without the presence of any foreground objects. However, foreground object detection is the process of extracting moving objects with prior knowledge of the background. Foreground object detection becomes more challenging in real-time environments for instance illumination variations, out of range scenarios and color camouflage of foreground objects concerning the background information \cite{maddalena2018background}. There are other challenging conditions as well, such as sequences with shadows or ghosting artifacts and also bootstrapping in which foreground objects are present in almost all frames of a video sequence. Over the past decade, many techniques have been proposed in the literature to address the problems of these challenging background scenes for the tasks of foreground detection and evaluation \cite{bouwmans2017scene, javed2017background, shimada2013background}.

To address the challenges mentioned above for object segmentation, we present a fusion based deep learning method. Our proposed method is based on Generative Adversarial Network (GAN) \cite {goodfellow2014generative} working on the idea of back propagation steps to generate specific kind of data. In this study, our primary focus is foreground object segmentation in the presence of various challenging conditions in background scenes. To address this problem, we present a solution based on GANs \cite{schlegl2017unsupervised} which works on the principle of generating background samples with scene specific information. Our proposed GAN model trains in an unsupervised manner on all video sequences irrespective of the presence of foreground objects in them as scene specific model. The primary purpose of training scene-specific GAN model is that the network should be able to learn the semantics of the scene containing various foreground background objects in the presence of challenging conditions such as illumination variations, shadows, and dynamic background information. The key idea behind training our GAN model with various challenging scenes is that during testing, our model will be able to generate the background scene according to our given test sample information. Since our network is trained on the data containing both background as well as foreground information, we need to eliminate moving foreground objects from our test sample in order to generate the exact background image via our trained GAN model. This can be done by multiplying test samples containing foreground objects with their motion masks evaluated by using optical flow \cite{liu2009beyond}. This step helps our GAN model to generate exact image sample containing only background information similar in semantics as test sample via back-propagation technique. The generated background sample is then subtracted from the given test sample to detect foreground objects as shown in Figure \ref{fig_framework}.

\section{Related Work}
In the last two decades, several methods have proposed for foreground detection by exploiting color and depth information. There are various traditional background subtraction methods, for instance, a very popular and classic method for background subtraction is GMM \cite{stauffer1999adaptive}. The idea of this proposed method is, modeling of each pixel is done with a mixture of Gaussian, and it works on color information processing only.
However, SOBS algorithms, such as \cite{maddalena2008self} and \cite{maddalena2012sobs}, shows improvement in performance by using color and depth features. They are based on self-organizing neural networks, which achieves better results in various challenging environments. Another efficient background subtraction model with a fusion of depth information embedded in the basic structure of \textit{Robust Principal Component Analysis} (RPCA) is proposed by Javed \textit{et al.} \cite {javed2017moving, javed2018moving}. It is a hybrid RPCA model with spatial and temporal information handling mechanism. Therefore, it is called \textit{Spatiotemporal Robust Principal Component Analysis} (SRPCA). SRPCA algorithm is a graph-regularized method, which preserves the spatiotemporal information of background that is low-rank matrix formation in the form of dual spectral graphs. To address the challenges in background subtraction problem,  Bo Xin \textit{et al.} \cite{xin2015background} presented a technique called \textit{Background Subtraction via Generalized Fused Lasso Foreground Modeling} (BS-GFL). Their primary objective is to address the problem of missing information in foreground detection because of various challenges like illumination variations and dynamic backgrounds. To solve this problem, they consider generalized fused lasso regularization to search for intact structured foregrounds to recover the missing content during the background subtraction process. Although BS-GFL is an efficient algorithm with a very good performance, however, it is an offline and partially supervised method. To solve the same problem of the missing content of foreground regions during background subtraction process, Xiaowei Zhou \textit{et al.} \cite{DECOLOR} proposed a method called \textit{Moving Object Detection by Detecting Contiguous Outliers in the Low-Rank Representation} (DECOLOR). The authors of this proposed method exploited Markov random field technique to recover the missing content of foreground detection during background subtraction process. For foreground detection by exploiting color as well as depth information, Massimo De Gregorio \textit{et al.} \cite{de2017cwisardh} presented an algorithm known as \textit{cwisardH+} working on the idea of decoupling the color information from the pixel depth information. The two video sequences runs are synchronously in this algorithm but independently and modeled by weightless neural networks at each pixel value.

\section{Proposed Method} \label{Proposed}
\begin{figure}[t]
	\centering
	\includegraphics[scale=0.28]{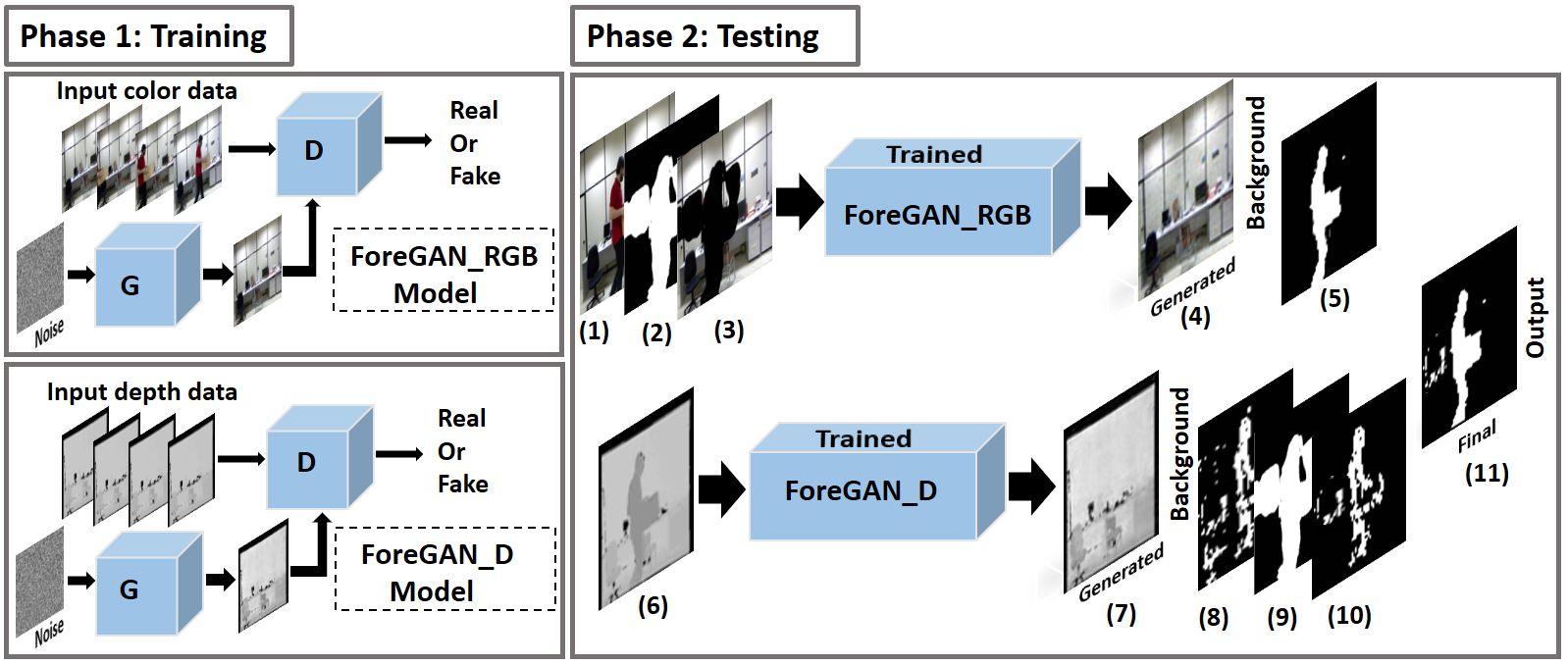}
	\caption{Workflow of the proposed ForeGAN\_RGBD model. Phase 1: Training has two independent models one to learn color information of different scenes, called ForeGAN\_RGB model and the other to learn depth information which is called ForeGAN\_D model. Phase 2: Testing also has two independent trained models for evaluation. (1) Input test RGB image sample, (2) Motion masks of test sample via optical flow, (3) Input RGB image sample multiplied by motion mask, (4) Generated RGB background image sample, (5) RGB foreground detection by equation \ref{FDL}, (6) Input test depth image sample, (7) Generated depth background image sample, (8) Depth foreground detection also by equation \ref{FDL}, (9) Motion mask, (10) Depth foreground detection after masking to eliminate noisy background pixels, and (11) Final output by fusion of results from (10) and (5) which is simply pixel by pixel addition. The samples presented in this Figure are from SBM-RGBD dataset \cite{camplani2017rgb} video sequence 'genSeq2' category 'Shadows'.}
	\label{fig_framework}
\end{figure}

In this section, we describe each step of our proposed algorithm in detail. The workflow diagram of our proposed method ForeGAN\_RGBD is presented in Figure \ref{fig_framework}. Our proposed method aims to perform foreground object segmentation by adding RGB and depth information into  DCGAN model \cite{radford2015unsupervised}. 
ForeGAN\_RGBD has two phases: Phase 1.) Training of the RGB and depth video sequences with two independent models as shown in Figure \ref{fig_framework}. Phase 2.) Testing of the two trained models with video sequences including foreground objects or background objects via backpropagation technique. It means that in phase 1 we learn two models based on GAN \cite{yeh2017semantic} representing indoor and outdoor scenes with various challenges on color as well as depth information. Each GAN train two models simultaneously, a discriminator model and a generator model, to differentiate between real and fake generated data containing RGB or depth information. The objective of the model is to achieve equilibrium of costs, increasing the ability of the model to generate data that is more accurate. In the following, we explain our proposed ForeGAN\_RGBD model in detail.
\subsection{ForeGAN\_RGBD Phase 1}\label{foregan}
A GAN model has two adversarial modules, a discriminator $D$ and a generator $G$. The main objective of generator $G$ is to learn a distribution $p_{gen}$ over input data $X_i$ via mapping of $z$ samples through $G(z)$. This mapping facilitates the $1D$ vectors of input noise which is uniformly distributed and sampled from latent space $Z$ to the $2D$ image representation. In a basic GAN model discriminator, $D$ is a Convolutional Neural Network (CNN) model that maps a $2D$ image representation to a single value $D(\cdot)$. This single value $D(\cdot)$ of discriminator's output can be explicated as the probability that whether the input given to the discriminator $D$ was a fake generated image $G(z)$ by the generator $G$ or a real image $X$ sampled from training data $X_i$. The discriminator and the generator are simultaneously optimized via cross entropy loss functions in a following two-player minimax game with $\Gamma (D, G)$ as a value function:
\begin{equation} \label{eq_fullgan}
\begin{split}
\min\limits_{G}\max\limits_{D}~\Gamma(G, D)= \mathbb{E}_{x\sim p_{data}(x)}[log(D(x))]\\
+ \mathbb{E}_{z\sim p_{z}(z)}[log(1 - D(G(z)))]. 
\end{split}
\end{equation}
The discriminator in GAN model is a decision maker entity which is trained to maximize the probability of assigning real training sample to actual input data and samples from $p_{gen}$ to the fake generated data. The purpose of the generator $G$ is to fool $D$ via minimizing the cost function $\Gamma(G)$ = $log(1 - D(G(z)))$, which is basically to maximize the following equation:
\begin{equation}
    \Gamma(G) = D(G(z)).
\end{equation}
During the training process, the generator tries to improve itself by generating realistic images, and the discriminator tries to identify the real and fake generated images. Once the training is done then the next step is testing which is discussed in detail in the next section. 
\subsection{Inverse Mapping of Test Images to Latent Space Representation, Phase 2}
During the phase of adversarial training, the generator learns the mapping from latent space representations $z$ which is random noise to more realistic images, $G(z) = z\mapsto x$. However inverse mapping $\theta(x) = x\mapsto z$ in GAN is not a straightforward process, rather we need a different mechanism for this purpose. To achieve inverse mapping Yeh \textit{et al.} \cite{yeh2017semantic} presented a back-propagation method to input data. This technique was already used by Google's DeepDream to create dreamlike images \cite{mordvintsev2015inceptionism}. Moreover, the back-propagation method has also been used to understand and visualize neural network's learned features by inverting the network by updating gradients at the level of input layer \cite{yeh2017semantic},  \cite{mahendran2015understanding}, \cite{linden1989inversion}, \cite{dosovitskiy2016inverting}. Every back-propagation based method requires specific loss functions for inverse mapping process in a neural network. Therefore our method has two loss functions as well. The purpose of the inverse mapping in our proposed method is to generate the specific kind of data during testing via back propagation strategy.
\subsubsection{Foreground Object Segmentation Loss} 
Given a test image $x$ we aim to find that particular random noise $z$ in the latent space that was mapped to generate image $G(z)$ via back-propagation method. To find that specific $z$, we have to select an initial random sample $z_o$, from the latent space and reinforce it to the trained generator network to generate $G(z_o)$. The loss functions are designed by the generated image $G(z_o)$, which provides significant update information of the coefficients of $z_o$ leading our back-propagation step to be shifted from $z_o$ to $z_1$ in the latent space distribution. The most similar generated image $G(z_{\beta})$ can be found by various back-propagation steps $\beta = 0, 1, 2, ... ,\phi$ by minimizing the following loss function:
\begin{equation}\label{FDL}
 \Upsilon_{F}(z_{\beta}) = \sum |x - G(z_{\beta})|,
\end{equation}
$\Upsilon_{F}(z_{\beta})$ function measures the visual dissimilarity between the test image $x$ and generated image $G(z_{\beta})$ via various back-propagation steps. In an ideal scenario if test and generated images are identical, $\Upsilon_{F}(z_{\beta})$ will be zero. 
\subsubsection{Feature Matching Loss} 
Now the next issue is how to generate those images $G(z_{\beta})$ which exactly makes the best match with test image $x$. To solve this problem, we need another loss term which helps the generator to generate similar images as test images and minimize the loss function in equation $\eqref{FDL}$. In contrast to the loss functions for best matching technique defined in \cite{yeh2017semantic}, Thomas Schlegl \textit{et al.} \cite{schlegl2017unsupervised} adopted an interesting method to lead generator to generate the specific images as test images. This method addresses the problem arises due to over-training of the discriminator, which causes the instability of GANs. To improve the inverse mapping of test image $x$ to that specific random noise $z$, feature matching technique is designed to force the generator to generate the data with similar statistics as test data. Since discriminator feeds its gradients to the generator, it is best to design the loss function on discriminator. This loss function is defined in such a way that the intermediate feature layer of the discriminator is feed with the generated image $G(z_{\beta})$:
\begin{equation} \label{ML}
    \Upsilon_{M}(z_{\beta}) = \sum |l(x) - l(G(z_{\beta}))|,
\end{equation}
Where $l(\cdot)$ represents the output of the intermediate layer of the discriminator, which describes the test image $x$. Based on this loss function, the discriminator is now used as a feature extractor rather than a decision maker for real or fake image representations during testing process only.\\
The overall loss functions can be represented as the weighted sum of both loss terms defined in equations \eqref{FDL} and \eqref{ML}:

\begin{equation} \label{final}
    \Upsilon_{M}(z_{\beta}) = (1 - \eta) \Upsilon_{F}(z_{\beta})+ \eta \Upsilon_{M}(z_{\beta}).
\end{equation}
The back-propagation method is only applied to the coefficients of $z$, while other hyper parameters of the trained GAN model remain unchanged in phase 2 which is Testing.

\subsection{Motion Masks via Optical Flow} \label{motion_mask_method}
To identify fast moving objects in video frames, we use optical flow \cite{liu2009beyond} which creates a motion mask to capture moving objects. This technique calculates motion between each pair of consecutive frames in the given input video sequence $I$. Motion mask $M$ is estimated by using motion information from consecutive frames of video sequences. Suppose $I_{t}$ and $I_{t-1}$ be the two consecutive video frames in $I$ at time any instant $t$ and $t-1$, respectively. Taking $U^{y}_{t,p}$ be the horizontal component and $V^{x}_{t,p}$ be the vertical component of the motion vector at position $p$ computed between consecutive frames, the corresponding motion mask, $M_{t}\in\{0,1\}$ will be estimated as:
\begin{equation} \label{eq_motion}
M_{t,p}=\begin{cases}
1,~~if~~\sqrt {(U^{y}_{t,p})^2 + (V^{x}_{t,p})^2 }< T,\\
0,~~otherwise. 
\end{cases}
\end{equation}
In the above equation, $T$ is a threshold of motion mask magnitude, and it is computed by taking the average of all pixels in the estimated motion field.
\section{Implementation} \label{implemention}
Our work is inspired by \cite{yeh2017semantic} and we have adapted the network \cite{schlegl2017unsupervised} which is based on DCGAN-tensorflow implementation \cite{radford2015unsupervised}. Our proposed ForeGAN\_RGBD model has a generative network, G, takes a random noise vector drawn from a uniform distribution and generates an image with dimension $64\times64\times3$. The discriminator model, however, runs in reverse order of the generator. The input images are fixed at $64\times64\times3$ running through a series of five convolution layers with down sampling and the number of channels are double the size of the previous convolution layer. Furthermore, the last layer of the network is a two class softmax. For training the ForeGAN\_RGBD model, we use Adam \cite{kinga2015method} for optimization. The back-propagation steps in our model are $2000$ during testing to generate the background image similar to the test image. The hyper-parameters are discussed in detail in section \ref{training} and \ref{testing}. Experiments are conducted using TitanX GPU with fixed input image samples of size in training as well as testing. Since all the categories have a limited number of video sequences, we have increased the training data by data augmentation technique which is translation and rotation of all the training images. Note that only those samples are considered in training whose ground truth information is not available in the dataset. Similarly only those video frames are considered for evaluation/testing purpose whose ground truth information is available in the dataset. 
\section{Experiments} \label{experiments}
We have evaluated our proposed approach on $6$ different categories of SBM-RGBD dataset \cite{camplani2017rgb} containing $27$ video sequences.
The video sequences are very challenging as they contain indoor and outdoor background scenes for foreground detection in the presence of color as well as depth information. The experimental process is divided into two phases, training and testing. 
The detail explanation of both phases is presented as follows:
\subsection{Phase 1: Training}\label{training}
The training is performed on $6$ categories of the SBM-RGBD dataset using the proposed technique.
We have trained two models independently one to learn of color information (\textit{ForeGAN\_RGB}) of the video sequences and the other to learn the depth information (\textit{ForeGAN\_D}) with $200$ epochs each.
During the training phase, the goal of our GAN network is to learn the color as well as depth information of the background scene. However, the only difference between two models is that the RGB model is trained on all kinds of video sequences regardless of the presence of foreground objects while the depth model of GAN is trained with the video sequences containing only depth information of background scene. Both of the models are trained individually on scene-specific information based on each category in SBM-RGBD dataset.
\begin{figure}[t]
	\centering
	\includegraphics[scale=0.32]{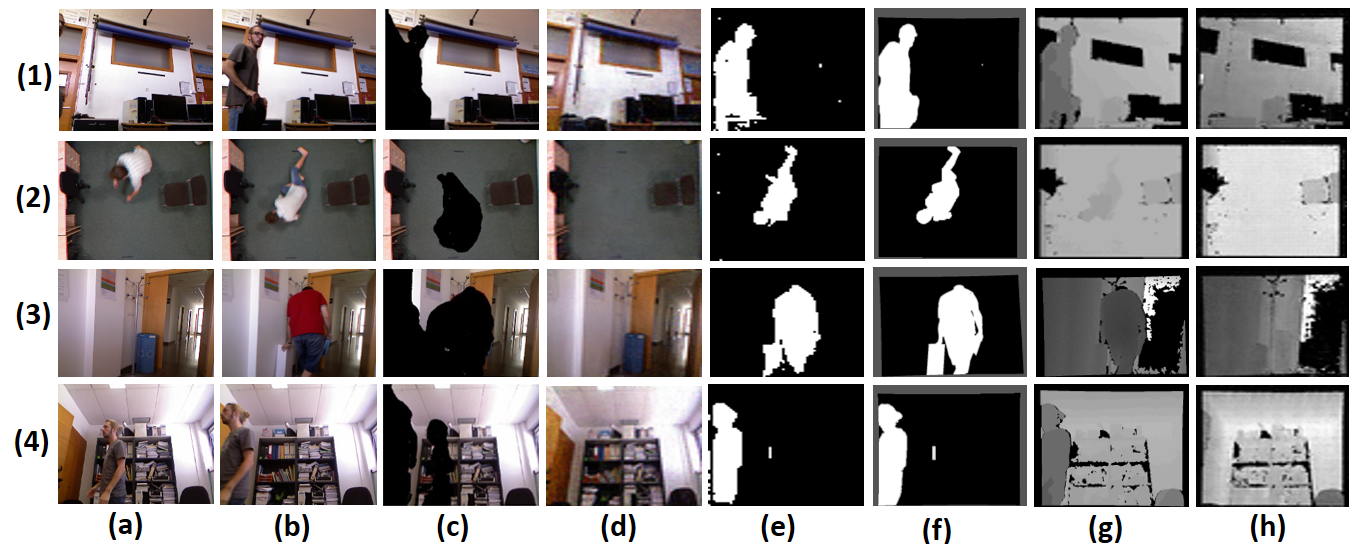}
	\caption{Visual results of each step of our propose ForeGAN\_RGBD method. (a) Input test RGB sequences $I_{t}$ as mentioned in section \ref{motion_mask_method} (b) Input test RGB sequences $I_{t-1}$ (c) Input test RGB sequence $I_{t-1}$ multiplied by motion masks to identify moving objects (d) generated background samples by our $ForeGAN\_RGB$ model (e) Foreground objects detected by subtracting (b) and (d) which is equation \eqref{FDL}, (f) Ground truths (g) Test depth sequences (h) Generated background depths by $ForeGAN\_D$ model. The sequences represented in $row~1$ is from 'Bootstrapping\_ds' category 'Bootstrapping', $row~2$ is from 'fall01cam1' category 'Shadows',  $row~3$ is from 'Hallway ' category 'Color Camouflage' and  $row~4$ is from 'Despatx\_ds' category 'Depth Camouflage'.}
	\label{stepby}
\end{figure}
\subsection{Phase 2: Testing}\label{testing}
The testing of the proposed models is performed independently on both trained models, and the output of both models is fused to get the final results.  All the testing samples are also fixed as mentioned in section \ref{implemention} and given as input to the models individually for validation with back-propagation steps. We have evaluated our proposed model by using $F$-measure score which is calculated as follows:
\begin{equation}
F =  2\frac{P\times R}{P + R},
\end{equation}
\begin{equation} \label{pre}
P =  \frac{T_{p}}{T_{p} + F_{p}},
\end{equation}
\begin{equation} \label{re}
R =  \frac{T_{p}}{T_{p} + F_{n}},
\end{equation}
where $T_{p}$ is True positives, $T_{n}$ is True negatives, $F_{p}$ is False positives, $F_{n}$ is False negatives, $F$ is $F$-Measure, $P$ is Precision, and $R$ is Recall.
For better foreground detection, the aim is to maximize $F$ score.
Since the two models, $ForeGAN\_RGB$ and $ForeGAN\_D$ are trained in way different way; they are also tested discretely, to improve the performance of the foreground object detection.
The trained model for color information $ForeGAN\_RGB$ is tested in such a way that the input test sequences are first multiplied with motion mask to eliminate the moving foreground object information (see Figure \ref{stepby} (3)).
This step helps our trained network to generate the background image sample with similar information by back-propagation steps.
Later this generated background image is subtracted with input test sample to extract the foreground objects (Figure \ref{stepby} (a), (c) and (d), Figure \ref{fig_framework} (5)).
The $ForeGAN\_D$ is also evaluated in the same way, but the only difference is that the testing sampling is given as it is to the trained model which generates the similar depth information as test sample but it will be a background scene.
After that pixel by pixel subtraction of generated background depth information and testing sample gives us foreground object segmentation with a lot of noise in it mostly as shown in Figure \ref{fig_framework} (8).
So motion masks are multiplied with foreground objects detected via depth information to eliminate the noisy background depth pixels as shown in Figure \ref{fig_framework} (8), (9) and (10) respectively.
The last step is to fuse the information of foregrounds detected by following fusion technique from both models to get the final output
$$
ForeGAN\_RGB + ForeGAN\_D = ForeGAN\_RGBD
$$
\begin{table}
\centering
\caption{Comparison of average $F$-measure score on SBM-RGBD dataset \cite{camplani2017rgb} concerning depth and color features. The first and second best performing methods are shown in red
and blue colors respectively.}
\scalebox{0.9}{
\label{f_table_rgbd}
\begin{tabular}{l*{6}{c}r}
\hline
Methods       & Depth features ~~~& RGB features  ~~~~&\\ 
\hline
BS-GFL         & 0.21      & 0.63 & \\ 
DECOLOR        &  0.20     & 0.56 &  \\ 
cwisardH+      & --- & --- & \\ 
RGB-SOBS        & --- & \textcolor{blue}{0.70} & \\ 
SRPCA           & \textcolor{blue}{0.24}  &  \textcolor{red}{0.71} & \\ 
ForeGAN\_RGBD   & \textcolor{red}{0.45}   & 0.51 & \\ 
\hline
\end{tabular}
}
\end{table}
\begin{table}
\centering
\caption{Comparison of average $F$-measure score on SBM-RGBD dataset \cite{camplani2017rgb}. Figure \ref{comp_fig} represents visual comparisons with other methods. The first and second best performing methods are shown in red and blue colors respectively.}
\scalebox{0.9}{
\label{f_table}
\begin{tabular}{l*{6}{c}r}
\hline
Categories       & BS-GFL & DECOLOR & cwisardH+ & RGB-SOBS & SRPCA & ForeGAN\_RGBD &\\
\hline
Illumination Changes  & 0.2991 & 0.4861 & \textcolor{blue}{0.4581} & 0.4527 & 0.4454 & \textcolor{red}{0.8158}\\
Color Camouflage      & 0.7333 & 0.7030 & \textcolor{blue}{0.9510} & 0.4864 & 0.8329 & \textcolor{red}{0.9635}\\
Depth Camouflage      & 0.7540 & 0.7252 & 0.7648 & \textcolor{blue}{0.8935} & 0.8083 & \textcolor{red}{0.9360}\\
Out of Range          & 0.7182 & 0.5874 & \textcolor{red}{0.8987} & 0.8527 & 0.8011 & \textcolor{blue}{0.8726}\\
Shadows               &0.5869 & 0.9051 & \textcolor{blue}{0.9264} & 0.9218 & 0.7591 & \textcolor{red}{0.9271} \\
Bootstrapping         & 0.5711 & 0.7601 & 0.5669 & 0.8007 & \textcolor{blue}{0.8098} & \textcolor{red}{0.8646} \\
\hline
Average               & 0.6104 & 0.6482 & 0.7609 & 0.7075  & \textcolor{blue}{0.7698} & \textcolor{red}{0.8966} \\
\hline
\end{tabular}
}
\end{table}
\subsection{Foreground Object Segmentation by ForeGAN\_RGBD} 
\begin{figure}[H]
	\includegraphics[scale=0.39]{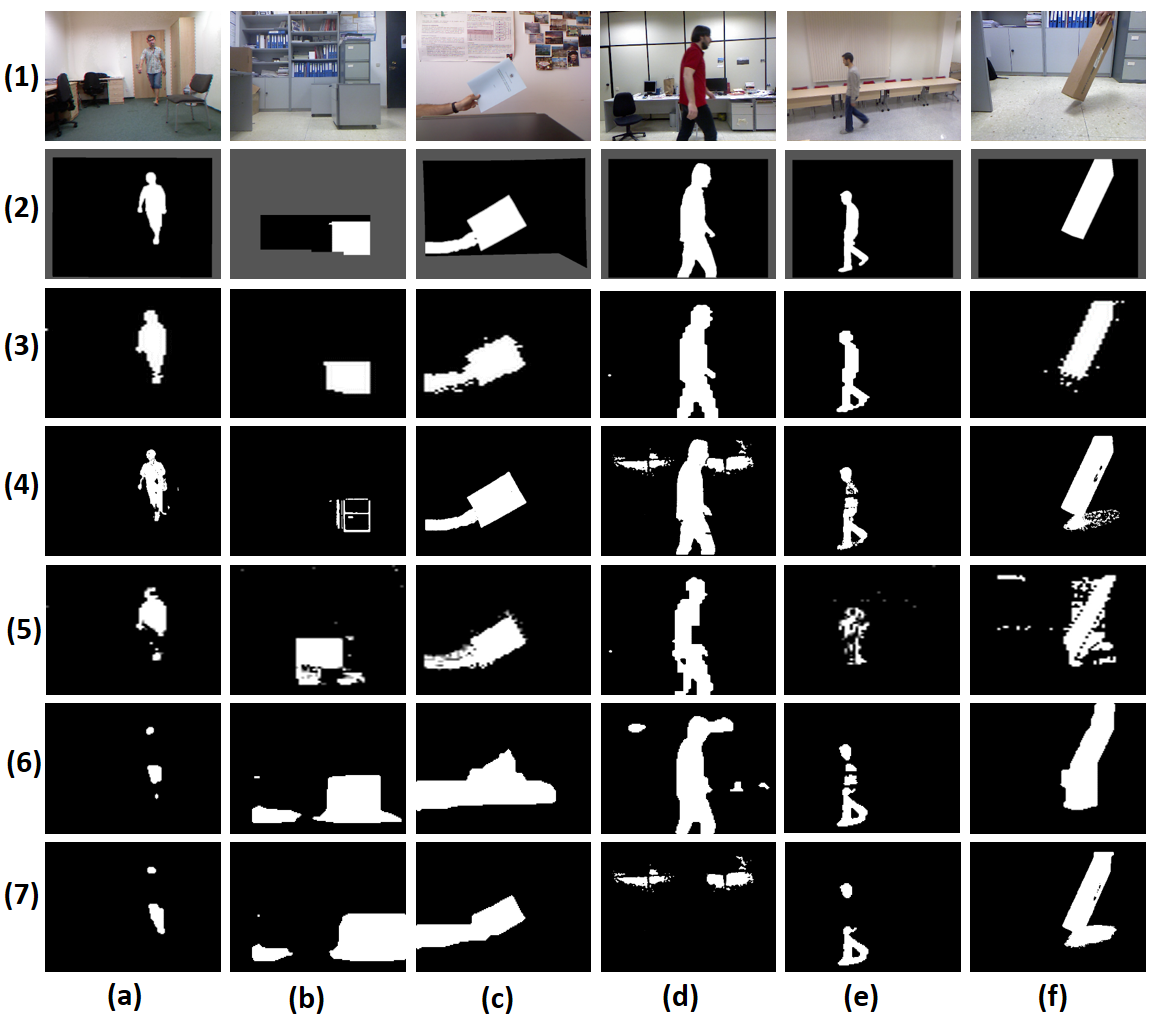}
	\caption{Performance comparison of our proposed method with 5 state-of-the-art methods.  From left to right: each input sequence is selected from different categories (a) 'fall01cam0' category 'Bootstrapping', (b) 'colorCam1' category 'Color Camouflage', (c) 'Wall' category 'Depth Camouflage', (d) 'genSeq1' category 'Illumination Changes', (e) 'MultiPeople1' category 'Out of Range' and (f) 'shadows1' category 'Shadows'. $row~1$: 6 images from input test video sequences $row~2$: Ground truth, $row~3$: foreground detection by our proposed ForeGAN\_RGBD method, $row~4$: SOBS-RGB, $row~5$: DECOLOR, $row~6$: SRPCA and $row~7$: BS-GFL.}
	\label{comp_fig}
\end{figure}

To highlight the significance of our proposed GAN model, we compared it with $5$ state-of-the-art methods by $F$-measure as shown in Table \ref{f_table}. By using original implementations of the authors, we have compared our proposed method with BS-GFL \cite{xin2015background}, DECOLOR \cite{DECOLOR}, cwisardH+ \cite{de2017cwisardh}, RGB-SOBS \cite{maddalena2012sobs} and SRPCA \cite{javed2017moving}. Qualitative comparisons of our proposed method are presented in Figure \ref{comp_fig} and $F$-measure score comparison with $5$ state-of-the-art methods are presented in Tables \ref{f_table_rgbd} and \ref{f_table}. It can be seen in Table \ref{f_table_rgbd} that our proposed method has performed well regarding foreground object segmentation by using depth features as compared to all methods. However, by using only RGB features the $F-measure$ is quite low which means our ForeGAN\_RGB model does not detect the foreground objects very well. The main reason behind this fact is that sometimes there are too much background noisy pixel values in frame difference detection of background-foreground images. Nevertheless, this problem is eliminated in ForeGAN\_D model because after frame difference of background depth image generated by our network and test sample; it is masked with optical flow motion information as shown in Figure \ref{fig_framework}. But upon combining the features of depth as well as color information the $F-measure$ is increased significantly in the case of our proposed algorithm as compared to all methods as shown in Table \ref{f_table_rgbd}. \\
The detail explanation of category wise comparison of our proposed method with $5$ state-of-the-art methods is as follows:\\
\textbf{Category: Illumination Changes} contains 4 video sequences in SBM-RGBD dataset \cite{camplani2017rgb}. The average $F$-measure scores among all four video sequences are shown in Table \ref{f_table}. It can be seen in Table \ref{f_table} that our proposed method ForeGAN\_RGBD has achieved maximum $F$-measure score among all the compared methods. However, the second best performing method is cwisardH+ among all the compared methods. Since this category represents challenging video sequences with several illumination variations it can be seen in Table \ref{f_table} that all compared methods except our proposed method have a very low $F$-measure score. It is because illumination variations pose a lot of challenges to all compared methods, but our proposed method can generate the background sequence with exact illumination condition. This aspect of our proposed method favors precise foreground detection with challenging illumination conditions. Visual results are represented in Figure \ref{comp_fig}.\\
\textbf{Category: Color Camouflage} also contains 4 video sequences including foreground objects that are very close in color to the background in SBM-RGBD dataset \cite{camplani2017rgb}. The average $F$-measure scores among all four video sequences are shown in Table \ref{f_table} and it can be seen that only our proposed method and cwisardH+ method has performed well. This category contains challenging video sequences in which foreground objects are very close in color to the background. Still, our proposed method achieved the highest $F$-measure score. The main reason is that as our algorithm extracts foreground objects from RGB and depth models independently, so if even one of the model suffers performance degradation the other model can detect the foreground objects. Qualitative results are shown in Figure \ref{stepby} and \ref{comp_fig} in comparison to ground truths and state-of-the-art methods respectively. \\
\textbf{Category: Depth Camouflage} also contains 4 videos including foreground objects very close in depth to the background in SBM-RGBD dataset \cite{camplani2017rgb}. It can be seen Table \ref{f_table} that our proposed method has outperformed all the compared methods and  RGB-SOBS method has achieved the second best score. The top performance of our proposed method is due to the same reason mentioned previously in color camouflage challenge discussion. The visual results are represented in Figure \ref{stepby} (d) row 4 and Figure \ref{comp_fig} (c). It can be seen in the qualitative analysis that our proposed method has better results than all the compared methods except RGB-SOBS method which has performed second best in this category. \\
\textbf{Category: Out of Range} has 5 videos sequences including foreground or background objects that are too close to/far from the sensor \cite{camplani2017rgb}. It can be seen in the Table \ref{f_table} that our proposed method has achieved second best $F$-measure score 0.8726 however cwisardH+ method achieved best score 0.8987. The reason is that sometimes our proposed method cannot generate the perfect background sequence for foreground objects which are too far from the sensor. Visual results presented in Figure \ref{comp_fig} (e) which shows the comparison of our proposed method with all compared methods and ground truth information as well.\\
\textbf{Category: Shadows} also contains 5 videos showing shadows caused by foreground objects. These can be visible-light shadows in the RGB channels or IR shadows in the depth channel \cite{camplani2017rgb}. In this challenging category, our proposed method has also achieved the highest $F$-measure score 0.9271 but the cwisardH+ method also achieved the best score with a minimal difference as shown in Table \ref{f_table}. The visual results shown in Figure \ref{stepby} (d) row 2 and Figure \ref{comp_fig} (f) represents that our proposed method can detect foreground objects even in the presence of shadows.\\
\textbf{Category: Bootstrapping} also contains 5 video sequences including foreground objects in almost all their frames \cite{camplani2017rgb}. It can be seen in Table \ref{f_table} that our proposed method ForeGAN\_RGBD has also achieved highest $F$-measure score and SRPCA has achieved second highest $F$-measure score. It is a very challenging category for all compared methods because it contains video sequences including foreground objects in almost all their frames but still our proposed method outperformed all compared methods. The main reason behind this fact is that our proposed method can learn the background scene information even if the foreground objects have left the scene once. This aspect of our proposed GAN network improves its efficiency in foreground object detection with challenging scenarios like bootstrapping. For instance visual results are shown in Figure \ref{stepby} (c) and (d) row 1 and Figure \ref{comp_fig} (a). 
\section{Conclusion}
In this study, we present the foreground detection algorithm based on generative adversarial network (GAN) exploiting color as well as depth information. Our goal is to do foreground segmentation in the presence of various major challenges in background scenes in real environments. To handle these challenges we present a fusion based foreground detection algorithm which exploits color as well as depth information, with the help of Generative Adversarial Network (GAN). Our proposed GAN model is trained in an unsupervised fashion on color and depth information independently with various challenges.  After that for testing, the GAN model has to generate the same background samples as test samples with similar statistics via back-propagation technique. The generated background samples are then subtracted from the given test samples to detect foreground objects from color as well as depth information, and final results are calculated by the fusion of both model's segmented outputs. The comparison of our proposed method with five state-of-the-art methods highlights the strength of our algorithm for foreground object segmentation in the presence of various challenging conditions.
\section{acknowledgements}
This research was supported by Development project of leading technology for future vehicle of the business of Daegu metropolitan city(No. 20181030).

\bibliographystyle{splncs04}
\bibliography{mybib}

\begin{thebibliography}{10}
\providecommand{\url}[1]{\texttt{#1}}
\providecommand{\urlprefix}{URL }
\providecommand{\doi}[1]{https://doi.org/#1}

\bibitem{bouwmans2018applications}
Bouwmans, T., Javed, S., Zhang, H., Lin, Z., Otazo, R.: On the applications of
  robust pca in image and video processing. Proceedings of the IEEE
  \textbf{106}(8),  1427--1457 (2018)

\bibitem{bouwmans2017scene}
Bouwmans, T., Maddalena, L., Petrosino, A.: Scene background initialization: a
  taxonomy. Pattern Recognition Letters  \textbf{96},  3--11 (2017)

\bibitem{bouwmans2014robust}
Bouwmans, T., Zahzah, E.H.: Robust pca via principal component pursuit: A
  review for a comparative evaluation in video surveillance. Computer Vision
  and Image Understanding  \textbf{122},  22--34 (2014)

\bibitem{camplani2017rgb}
Camplani, M., Maddalena, L., Gabriel, M., Petrosino, A., Salgado, L.: Rgb-d
  dataset: Background learning for detection and tracking from rgbd videos. In:
  IEEE ICIAP-Workshops (2017)

\bibitem{chen2018scom}
Chen, Y., Zou, W., Tang, Y., Li, X., Xu, C., Komodakis, N.: Scom:
  Spatiotemporal constrained optimization for salient object detection. IEEE
  Transactions on Image Processing  \textbf{27}(7),  3345--3357 (2018)

\bibitem{de2017cwisardh}
De~Gregorio, M., Giordano, M.: Cwisardh+: Background detection in rgbd videos
  by learning of weightless neural networks. In: International Conference on
  Image Analysis and Processing. pp. 242--253. Springer (2017)

\bibitem{dosovitskiy2016inverting}
Dosovitskiy, A., Brox, T.: Inverting visual representations with convolutional
  networks. In: Proceedings of the IEEE Conference on Computer Vision and
  Pattern Recognition. pp. 4829--4837 (2016)

\bibitem{goodfellow2014generative}
Goodfellow, I., Pouget-Abadie, J., Mirza, M., Xu, B., Warde-Farley, D., Ozair,
  S., Courville, A., Bengio, Y.: Generative adversarial nets. In: Advances in
  neural information processing systems. pp. 2672--2680 (2014)

\bibitem{javed2017moving}
Javed, S., Bouwmans, T., Sultana, M., Jung, S.K.: Moving object detection on
  rgb-d videos using graph regularized spatiotemporal rpca. In: International
  Conference on Image Analysis and Processing. pp. 230--241. Springer (2017)

\bibitem{javed2018moving}
Javed, S., Mahmood, A., Al-Maadeed, S., Bouwmans, T., Jung, S.K.: Moving object
  detection in complex scene using spatiotemporal structured-sparse rpca. IEEE
  Transactions on Image Processing  (2018)

\bibitem{javed2016spatiotemporal}
Javed, S., Mahmood, A., Bouwmans, T., Jung, S.K.: Spatiotemporal low-rank
  modeling for complex scene background initialization. IEEE Transactions on
  Circuits and Systems for Video Technology  (2016)

\bibitem{javed2017background}
Javed, S., Mahmood, A., Bouwmans, T., Jung, S.K.: Background--foreground
  modeling based on spatiotemporal sparse subspace clustering. IEEE
  Transactions on Image Processing  \textbf{26}(12),  5840--5854 (2017)

\bibitem{kinga2015method}
Kinga, D., Adam, J.B.: A method for stochastic optimization. In: International
  Conference on Learning Representations (ICLR). vol.~5 (2015)

\bibitem{linden1989inversion}
Linden, A., Kindermann, J.: Inversion of multilayer nets. In: Proc. Int. Joint
  Conf. Neural Networks. vol.~2, pp. 425--430 (1989)

\bibitem{liu2009beyond}
Liu, C., et~al.: Beyond pixels: exploring new representations and applications
  for motion analysis. Ph.D. thesis, Massachusetts Institute of Technology
  (2009)

\bibitem{maddalena2012sobs}
Maddalena, L., Petrosino, A.: The sobs algorithm: what are the limits? In:
  Computer Vision and Pattern Recognition Workshops (CVPRW), 2012 IEEE Computer
  Society Conference on. pp. 21--26. IEEE (2012)

\bibitem{maddalena2018background}
Maddalena, L., Petrosino, A.: Background subtraction for moving object
  detection in rgbd data: A survey. Journal of Imaging  \textbf{4}(5), ~71
  (2018)

\bibitem{maddalena2008self}
Maddalena, L., Petrosino, A., et~al.: A self-organizing approach to background
  subtraction for visual surveillance applications. IEEE Transactions on Image
  Processing  \textbf{17}(7), ~1168 (2008)

\bibitem{mahendran2015understanding}
Mahendran, A., Vedaldi, A.: Understanding deep image representations by
  inverting them. In: Proceedings of the IEEE conference on computer vision and
  pattern recognition. pp. 5188--5196 (2015)

\bibitem{mordvintsev2015inceptionism}
Mordvintsev, A., Olah, C., Tyka, M.: Inceptionism: Going deeper into neural
  networks. Google Research Blog. Retrieved June  \textbf{20}(14), ~5 (2015)

\bibitem{radford2015unsupervised}
Radford, A., Metz, L., Chintala, S.: Unsupervised representation learning with
  deep convolutional generative adversarial networks. arXiv preprint
  arXiv:1511.06434  (2015)

\bibitem{schlegl2017unsupervised}
Schlegl, T., Seeb{\"o}ck, P., Waldstein, S.M., Schmidt-Erfurth, U., Langs, G.:
  Unsupervised anomaly detection with generative adversarial networks to guide
  marker discovery. In: International Conference on Information Processing in
  Medical Imaging. pp. 146--157. Springer (2017)

\bibitem{shimada2013background}
Shimada, A., Nagahara, H., Taniguchi, R.i.: Background modeling based on
  bidirectional analysis. In: Computer Vision and Pattern Recognition (CVPR),
  2013 IEEE Conference on. pp. 1979--1986. IEEE (2013)

\bibitem{stauffer1999adaptive}
Stauffer, C., Grimson, W.E.L.: Adaptive background mixture models for real-time
  tracking. In: Computer Vision and Pattern Recognition, 1999. IEEE Computer
  Society Conference on. vol.~2, pp. 246--252. IEEE (1999)

\bibitem{sultana2018unsupervised}
Sultana, M., Mahmood, A., Javed, S., Jung, S.K.: Unsupervised deep context
  prediction for background foreground separation. arXiv preprint
  arXiv:1805.07903  (2018)

\bibitem{vaswani2017robust}
Vaswani, N., Bouwmans, T., Javed, S., Narayanamurthy, P.: Robust pca and robust
  subspace tracking. arXiv preprint arXiv:1711.09492  (2017)

\bibitem{wu2017moving}
Wu, Y., He, X., Nguyen, T.Q.: Moving object detection with a freely moving
  camera via background motion subtraction. IEEE Transactions on Circuits and
  Systems for Video Technology  \textbf{27}(2),  236--248 (2017)

\bibitem{xin2015background}
Xin, B., Tian, Y., Wang, Y., Gao, W.: Background subtraction via generalized
  fused lasso foreground modeling. In: Proceedings of the IEEE Conference on
  Computer Vision and Pattern Recognition. pp. 4676--4684 (2015)

\bibitem{yeh2017semantic}
Yeh, R.A., Chen, C., Lim, T.Y., Schwing, A.G., Hasegawa-Johnson, M., Do, M.N.:
  Semantic image inpainting with deep generative models. In: Proceedings of the
  IEEE Conference on Computer Vision and Pattern Recognition. pp. 5485--5493
  (2017)

\bibitem{zhang2015robust}
Zhang, T., Liu, S., Ahuja, N., Yang, M.H., Ghanem, B.: Robust visual tracking
  via consistent low-rank sparse learning. International Journal of Computer
  Vision  \textbf{111}(2),  171--190 (2015)

\bibitem{DECOLOR}
Zhou, X., Yang, C., Yu, W.: Moving object detection by detecting contiguous
  outliers in the low-rank representation. IEEE T-PAMI  \textbf{35}(3),
  597--610 (2013)

\end{thebibliography}
%
%
\end{document}